\documentclass[11pt]{article}
\usepackage{coling2020}
\usepackage{times}
\usepackage{url}
\usepackage{latexsym}
\usepackage{multirow}
\usepackage{subcaption}
\usepackage{graphicx}
\usepackage{caption}

\colingfinalcopy 

\title{A simple method for domain adaptation of sentence embeddings}

\author{Anna Kruspe \\
  German Aerospace Center (DLR) \\
  Institute of Data Science \\
  Jena, Germany \\
  {\tt anna.kruspe@dlr.de}
  }

\date{}

\begin{document}
\maketitle
\begin{abstract}
Pre-trained sentence embeddings have been shown to be very useful for a variety of NLP tasks. Due to the fact that training such embeddings requires a large amount of data, they are commonly trained on a variety of text data. An adaptation to specific domains could improve results in many cases, but such a finetuning is usually problem-dependent and poses the risk of over-adapting to the data used for adaptation.\\
In this paper, we present a simple universal method for finetuning Google's Universal Sentence Encoder (USE) using a Siamese architecture. We demonstrate how to use this approach for a variety of data sets and present results on different data sets representing similar problems. The approach is also compared to traditional finetuning on these data sets. As a further advantage, the approach can be used for combining data sets with different annotations. We also present an embedding finetuned on all data sets in parallel.
\end{abstract}

\section{Introduction}
\label{intro}
Deep learning technologies for text data started advancing rapidly with the introduction of word embeddings such as \textit{word2vec} \cite{mikolov}. Recently, two main advances were introduced into these technologies: Taking a broader context into account, such as in \textit{BERT} and related approaches \cite{bert}, and embedding whole sentences rather than words. Examples of such sentence embeddings include the Smooth Inverse Frequency (SIF) \cite{arora}, the Mean-Max Attention Autoencoder (Mean-Max AAE) \cite{zhang}, and the Universal Sentence Encoder (USE) \cite{cer}. For the latter two methods, a dedicated training on large text corpora is necessary. Pre-trained models are available online. Due to the requirement for a large amount of training data as well as a desire for wide applicability, these models are usually not focused on specific types of text or topics. Such general-purpose models may not be the best choice for more restricted problems, however.\\
It is not immediately obvious how to fine-tune these models to domains or genres of text. When annotated data is available, a naive approach consists of constructing a full model (e.g. a classification model) for the task, and allowing the embedding layers' weights to be adapted in the training procedure. This has several disadvantages. First of all, constructing a specific model for every task is necessary. This also means that a model cannot be easily adapted to several tasks in parallel. Additionally, such an approach is prone to over-adapting to the classes provided as opposed to the general domain (e.g. overfitting to events shown in training rather than the general task of event classification).\\
In this paper, we present an easier method for finetuning the Universal Sentence Encoder. Our approach utilizes a Siamese architecture. We demonstrate adaptation on several Twitter data sets designed for tasks such as (crisis) event detection, topic classification, or sentiment recognition. We also show how to finetune the embedding to all tasks in parallel.

%
%
\blfootnote{
    %
    %
    %
    %
    %
    %
}


\section{Proposed method}
\subsection{Sentence embeddings}
The development of semantic embeddings of words and sentences is an area of deep learning that has moved forward quickly since the introduction of the \textit{word2vec} algorithm in 2013 \cite{mikolov}. The idea behind these embeddings is a transformation into a latent space, i.e. a projection onto high-dimensional vectors that allow for numeric computations. Such representations are frequently used as the inputs to a continuative neural network, e.g. for classification tasks, instead of a mere one-hot encoding (i.e. a binary representation of word presence that does not contain any semantic information). Embedding models themselves are also neural networks, and are commonly trained on the assumption that words appearing in similar contexts are going to be linguistically similar.\\
Embedding models have commonly been trained on a word basis, but sentence-level models have become more popular in recent years. One such model is Google's Universal Sentence Encoder (USE) published by Cer et al. \cite{cer}. They show two versions, one using a Transformer structure and one using Deep Averaging Networks, and train them on a large set of combined data sources. These sources are Wikipedia, web news, web question-answer pages and discussion forums, augmentend with unsupervised learning on the Stanford Natural Language Inference (SNLI) corpus \cite{bowman}. The Transformer version is available from the authors on TensorFlow Hub\footnote{\url{https://tfhub.dev/google/universal-sentence-encoder-large/5}}. For our experiments, we use the Transformer version of USE on a tweet-level basis (i.e. we treat each tweet as a sentence for the purposes of this model, even though this is not always the case). This leads to a 512-dimensional representation of each tweet.

\subsection{Related work}
Considering that Twitter data was not part of the training data, we expect these embeddings to produce good, but not optimal results for tasks on this kind of data. Finetuning them to different kinds of data or, viewed differently, adapting them to other domains, could improve the latent representation for these purposes. Finetuning word embeddings has been a topic of research since their inception. Original and adapted embeddings are often concatenated. Methods mainly rely on either training them from scratch on domain-specific data (e.g. \cite{li}) or projecting word embeddings into a new latent space (e.g. \cite{sarma,faruqui,barnes}).  The latter can, for example, be implemented with autoencoders, Canonical Correlation Analysis (CCA), or maximum a posteriori (MAP) estimation, either within a training process or separately. These methods require domain-specific training data that defines, either implicitly or explicitly, similarities between certain words (e.g. through a lexicon).\\
Both training from scratch and projection approaches cannot be easily transferred from word to sentence embeddings. Due to the complexity of these models, a large amount of training data is required for training, preferably for multiple tasks within one domain. This amount of data is usually not available for a particular domain. Projection cannot be used due to the  fact that similarity measures are also much more difficult to define for sentences than for words. In addition, word embeddings usually cover a finite number of entries, whereas there is an infinite number of possible sentence embeddings for a given encoder.\\

\subsection{Naive adaptation}\label{subsec:approach_naive}
\begin{figure}[htbp]
\centerline{\includegraphics[width=.4\textwidth]{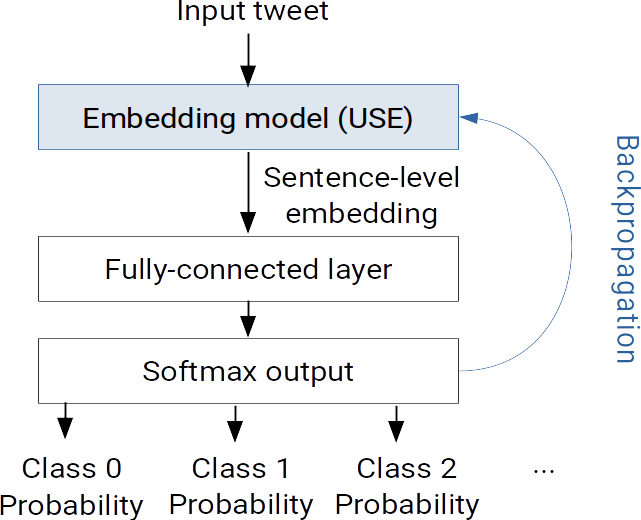}}
\caption{Architecture for naive finetuning approach.}
\label{fig:finetune_naive}
\end{figure}

Consequently, there is no standard method for finetuning sentence embeddings. The most obvious method consists of constructing a model that comprises the pre-trained embedding model at the start, followed by feed-forward layers leading to an output appropriate for the problem at hand. A schematic of such an architecture is shown in figure \ref{fig:finetune_naive} (our fully-connected layer has 128 dimensions and ReLU activations). The embedding model's layers are allowed to be adapted through backpropagation. Once training has converged, the later layers can be discarded and embedding model finetuned in this manner can be used for other tasks.\\
This approach generally works, but has several disadvantages. First of all, the model will likely become highly adapted to the exact classes given in the data, rather than the domain in question. As an example, when finetuning a model for event detection, the embedding will learn to distinguish the events given in the training data, but will not necessarily generalize to new events. Beyond this class-level overfitting, there is also a risk of overfitting to the training data itself, especially for small data sets.\\
A second difficulty arises when multiple data sets are combined for training. Even when those data sets represent the same domain, labels may be given in a different format or on different scales. This makes conversion necessary, which may introduce artifacts into the training data. The problem becomes much graver when combining data sets from multiple domains to create an embedding that can serve different purposes at the same time, or that is expected to cover a more general type of text (e.g. ``Twitter data'' or ``Disaster tweets'') rather than a specific task. Such models can be trained with a multi-task architecture, but preliminary experiments show that it is difficult to balance the model among the various tasks.

\subsection{Siamese adaptation}\label{subsec:approach_siamese}
\begin{figure}[htbp]
\centerline{\includegraphics[width=.7\textwidth]{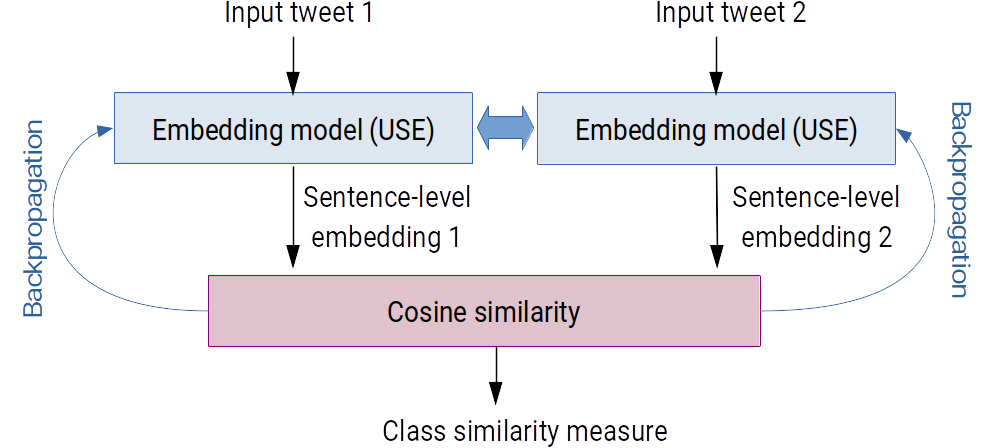}}
\caption{Architecture for Siamese finetuning approach.}
\label{fig:finetune_siamese}
\end{figure}

For these reasons, we propose a different way of finetuning sentence embeddings that is inspired by one-shot learning techniques. Namely, we finetune the embedding with a Siamese architecture. This model consists of two identical branches both containing the embedding with weights shared between them. Two examples (in our case tweets) at a time are given as the input to the model. These examples either belong to the same class defined in the training labels, or to different ones. The inputs are transformed to the 512-dimensional latent space with the model; the cosine similarity between both is the model's output. This architecture is shown in figure \ref{fig:finetune_siamese}. Training labels are binary, stating whether the inputs belong to the same class or not. In consequence, this is what the model learns. What this means for the embedding is that the latent space adapts to the domain in question, and learns in what way the classes differ.\\
This makes the training process very flexible. So-called ``episodes'' of two tweets and their binary label can be generated dynamically from any data set with class annotations. This means that different data sets can be combined very easily, both for the same domain and for different ones. To balance different tasks, the same amount of episodes can be generated for each of them. In our experiments, we show how to perform siamese adaptation both for single tasks as well as multiple ones at the same time.

\subsection{Evaluation}\label{subsec:approach_eval}
The measure of quality of an embedding for a certain classification task is its ability to distiguish between the classes in question. We therefore calculate the cosine distance between examples in our test data sets. This could be done for each combination. We simplify the process by once again generating random pairs of tweets of the same or different classes, calculating the distance between them, and then taking the average distance for same and different classes:
\begin{equation}
 \Delta {Cosine Distance} = \frac{\sum_{d=0..D} C(p_d)}{D} - \frac{\sum_{s=0..S} C(q_s)}{S}
\end{equation} 
where $p_d$ are the random different-class example pairs, $q_s$ are the random same-class example pairs, and $C(r_t)$ is the Cosine difference of a pair $r_t$ with vectors $r_{t0}$ and $r_{t1}$:
\begin{equation}
 C(r_t) = 1 - \frac{r_{t0} r_{t1}}{||r_{t0}||_2 ||r_{t1}||_2}
\end{equation} 

\begin{figure}[htbp]
    \centering
    \begin{subfigure}[b]{0.5\textwidth}
        \centering
        \includegraphics[width=.9\textwidth]{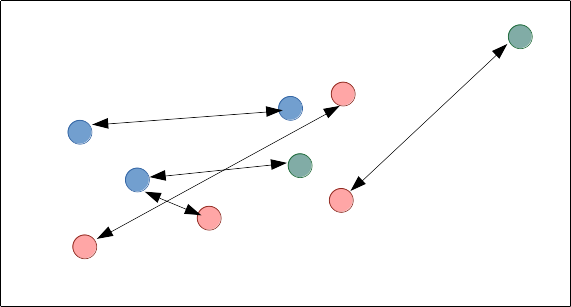}
        \caption{Bad example}
    \end{subfigure}%
    \begin{subfigure}[b]{0.5\textwidth}
        \centering
        \includegraphics[width=.9\textwidth]{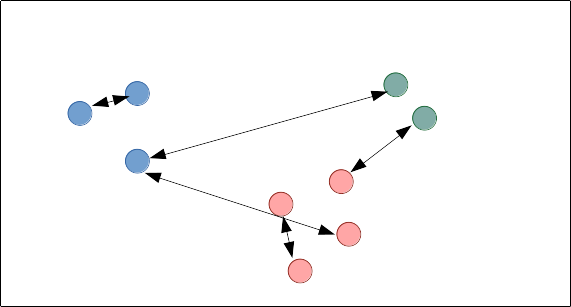}
        \caption{Good example}
    \end{subfigure}%
    \caption{Two examples of random evaluated pairs of tweet embeddings and their distances in 2D; colors represent labelled classes. In the bad example, distances of pairs with the same class label are not smaller than of those with different classes. In the good example, they are.}
    \label{fig:eval_example}
\end{figure}

The difference between these average distances indicates the embedding's decision performance. Using random examples has two advantages: First, it speeds up the evaluation; second, it allows us to ensure that the same number of test examples are used for each class and task. Two examples are shown in figure \ref{fig:eval_example}.

\section{Data sets}
In order to evaluate the versatility of our approach, we test it on nine data sets. All of them consist of data collected from Twitter (tweets) for a variety of tasks. The usage of Twitter data is motivated by two points: First, we wanted to test the approach on a domain that was not present in the original training for USE. In this way, we can also attempt to finetune a model to this type of text rather then the particular tasks. Second, we want the experiments for each data set to be comparable and also be able to flexibly exchange training and test data sets (e.g. test models on a different domain than the one they were trained on). Seven of the data sets are used for training; some of them also contain test sets that were used for validation, and for some, we chipped of a test set in advance. The remaining two were used exclusively for testing.\\
\begin{description}
\item[CrisisNLP] The team behind \textit{CrisisNLP} collected tweets during 19 natural and health-related disasters and published them for research \cite{imran}. Collected tweets range between 17,000 and 28 million per event, making up around 53 million in total. Out of these, around 50,000 were annotated both by volunteers and by paid workers on the \textit{CrowdFlower} crowdsourcing platform (now named \textit{Figure Eight}\footnote{\url{https://www.figure-eight.com/}}  with regard to information type. In this work, only the tweets with \textit{CrowdFlower} annotations were utilized. These tweets come from a subset of 11 English-language event sets, summing up to around 23,000 in total. We only utilize the event annotations, not the semantic classes. This data set is only used for training.
\item[Events2012] This data set contains 120 million tweets, of which around 150,000 were labeled to belong to one of 501 events (which are often better interpretable as topics) \cite{mcminn}. We use a subset of 90,000 of these tweets as not all original ones are still available. We randomly split this data set by events to produce a training set (401 events) and a test set (100 events).
\item[Gender] This data set was collected by \textit{Figure Eight} themselves, and then annotated by volunteers on the platform\footnote{\url{https://www.figure-eight.com/data-for-everyone/}}. They were asked to judge the gender of each tweet's author according to the categories ``male'', ``female'', ``brand or organization'', or ``can't tell''. \textit{Figure Eight} then performed various classification experiments and analyses\footnote{\url{https://www.figure-eight.com/using-machine-learning-to-predict-gender/}}. Among other results, they state that their classifier was not highly confident, but was still able to detect certain patterns; this is usually the case for this sort of task \cite{lopesfilho}. The whole data set consists of around 25,000 tweets, which we randomly split into a training set of 20,000 tweets and a test set of 5,000 tweets.
\item[Flu] Presented in \cite{lamb}, this data set contains a collection of tweets concerning seasonal influenza. Tweets were collected by keywords during two flu seasons (2009 and 2012), according to three concepts: Related vs. not related, self vs. others, and awareness vs. infection. Annotation was performed by paid Amazon Mechanical Turk\footnote{\url{http://www.mturk.com}} workers. In this work, we only use data of the first concept. The ``not related'' category contains tweets that do not deal with the seasonal flu, but could easily be confused because they contain similar keywords (e.g. ``flu'' in the context of swine flu). There are around 2,300 tweets for 2009, and around 4,700 for 2012. Due to the higher amount of 2012 tweets, we decided to use that year as training data, and the 2009 data as our test set to evaluate transferability.
\item[Political] \cite{marchetti} provide a data set of around 2,000 tweets dealing with political topics, and another 2,000 ones about other topics for comparison. These tweets were partially obtained by searching for pre-defined keywords, and then annotated by the authors. We split the data set randomly into a set of 3,200 training tweets and 800 test tweets.
\item[SemEval-2015 PIT] This data set was originally provided for the 2015 SemEval Paraphrase and Semantic Similarity In Twitter (PIT) challenge \cite{semeval}. Its training partition consists of ~13,000 tweets collected according to the trending topics on Twitter, and then annotated on Mechanical Turk with respect to their similarity (including paraphrases). Dev and test sets are also provided, which we combine into a new test set of around 5,700 tweets. We do not utilize the similarity annotations in this work, instead using this corpus for topic classification.
\item[Sentiment140] This is a data set developed for sentiment detection in tweets \cite{go}. The tweets were collected through keyword search, and then annotated automatically by detecting emoji; tweets are determined to have positive, neutral, or negative sentiment. The whole data set contains 1.6 million tweets; to reduce processing times, we only use a random subset of 50,000 for training. A test set of 500 tweets is also provided and used in this work. 
\item[CrisisLexT26] \textit{CrisisLex} was first published by Olteanu et al. in 2014 \cite{olteanu} and expanded later to \textit{CrisisLexT26} \cite{olteanu2}. It contains tweets collected during 26 crises, mainly natural disasters like earthquakes, wildfires and floods, but also human-induced disasters like shootings and a train crash. Amounts of these tweets per disaster range between 1,100 and 157,500. In total, around 285,000 tweets were collected. They were then annotated by paid workers on \textit{CrowdFlower} according to three concepts: Informativeness, information type, and tweet source. There is no overlap with the events in other data sets. This data set is only used for testing.
\item[Sanders] This data set was originally provided by Sanders Analytics\footnote{\url{https://web.archive.org/web/20180303104713/http://www.sananalytics.com/lab/twitter-sentiment/}, now available under \url{https://github.com/zfz/twitter_corpus}}. It consists of ~5,500 tweets that were manually annotated both with a technological topic (``apple'', ``google'', ``microsoft'', or ``twitter'') and a sentiment (positive/negative/neutral/irrelevant). We only use this data set for testing in both domains.
\
\end{description}
Table \ref{tab:configs} shows the combinations of training and test data we use in the following experiments, along with an interpretation of the semantic purpose of the finetuned models.
\begin{table}[]
  \centering
\begin{tabular}{ | l | l | l | l |} \hline 
  \textbf{Experiment} & \textbf{Training set} & \textbf{Test sets} & \textbf{Purpose} \\ \hline
  \hline
  A & CrisisNLP & CrisisLexT26 & Classification of crisis events \\ \hline
  \multirow{2}{*}{B} & \multirow{2}{*}{Events2012 (train)} & Events2012 (test) & \multirow{2}{*}{Classification of events/topics}\\
    & &SemEvalTest &\\ \hline
   \multirow{2}{*}{C} & \multirow{2}{*}{SemEval}&SemEvalTest&\multirow{2}{*}{Classification of topics}\\
    & &Sanders (topics)&\\  \hline
    \multirow{2}{*}{D} & \multirow{2}{*}{Sentiment140}&Sentiment140Test & \multirow{2}{*}{Sentiment recognition} \\
   &  &Sanders (sentiments) &\\ \hline
   E & Gender (train) & Gender (test) & Recognition of author's gender \\ \hline
   F & Flu2012 & Flu2009 & Detection of tweets about the seasonal flu (binary) \\ \hline
   G & Political (train) & Political (test) & Detection of tweets with political content (binary) \\ \hline
   
\end{tabular}
\caption{Train/test configurations of data sets used in our experiments.} \label{tab:configs}
\end{table}

\section{Experimental validation}
As shown in table \ref{tab:configs}, we train models for each of our training data sets, then test those models on one or two unseen test sets. This is done for seven configurations covering different tasks. In each of these experiments, we try four models:
\begin{itemize}
 \item The unmodified USE model (ORIG)
 \item A model naively finetuned on the data set as described in section \ref{subsec:approach_naive} (NAIVE)
 \item A model finetuned with the Siamese approach described in section  \ref{subsec:approach_siamese} (SIAMESE)
 \item A model trained on all of the training sets in parallel with the Siamese architecture as described in section \ref{subsec:approach_siamese} (ALL)
\end{itemize}
This means we evaluate the following embedding models in total: The original USE, a naively and a Siamese-finetuned model per experiment, and a Siamese model finetuned on all data sets at once. The ORIG model is not trained, but evaluated on the test sets from table \ref{tab:configs} as a baseline. The NAIVE and SIAMESE models are trained and tested on the combinations in that table. The ALL model is trained once on example pairs from all seven training data sets in parallel, and evaluated on the test sets from table \ref{tab:configs} individually.\\
Evaluation is performed as described in section \ref{subsec:approach_eval}: By calculating the difference in Cosine distance between tweets from the same class and tweets from different classes of the test set. Due to the fact that training labels are binary, distance values of pairs do not have an upper bound, but are commonly in the range between 0 and 1 (the higher, the better).\\
For the NAIVE models, we use the entire training data set in each experiment. For the SIAMESE finetuning, we randomly generate 70,000 example pairs from each experiment's training set.
For the ALL model trained on all datasets, 10,000 pairs per training set are generated, summing up to 70,000 in total as well. Each training is performed for 30 epochs. Evaluation is performed on 5,000 randomly generated pairs of the considered test set.

\begin{figure}[htbp]
    \centering
    \begin{subfigure}[b]{0.4\textwidth}
        \centering
        \includegraphics[width=.9\textwidth]{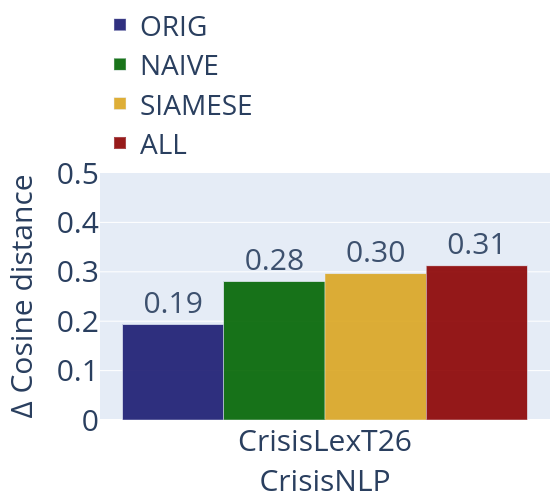}
        \caption{Experiment A}\label{fig:resA}
    \end{subfigure}%
    \begin{subfigure}[b]{0.6\textwidth}
        \centering
        \includegraphics[width=.9\textwidth]{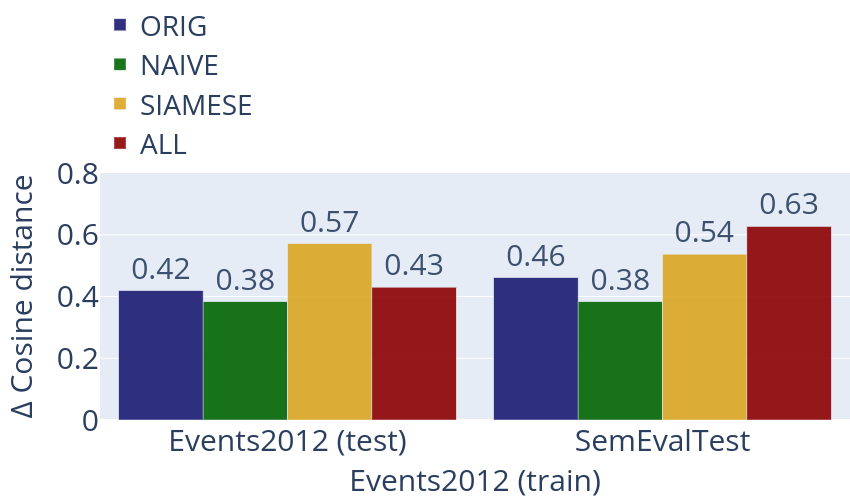}
        \caption{Experiment B}\label{fig:resB}
    \end{subfigure}%
    \caption{Experimental results}
    \label{fig:resAB}
\end{figure}
\subsection{Experiment A: CrisisNLP $\rightarrow$ CrisisLexT26}
Figure \ref{fig:resA} shows the results for all four models on our first train/test combination: \textit{CrisisNLP} and \textit{CrisisLexT26}. This finetuning serves to better distinguish between tweets concerning crisis events. The naive approach, where the model is finetuned based on the classes defined in \textit{CrisisNLP}, shows a clear improvement over the original USE baseline. The novel Siamese approach is slightly better; we believe that, although the effect is not large, the naive model may become somewhat focused on \textit{CrisisNLP}'s specific events. The embedding finetuned on all training sets is slightly better than the Siamese one. The fact that other training sets deal with similar tasks may be of help here.

\subsection{Experiment B: Events2012 (train) $\rightarrow$ Events2012 (test) \& SemEvalTest}
Results for the second experiment are shown in figure \ref{fig:resB}. The naive and Siamese models were finetuned on the train set of \textit{Events2012}, and all of the embeddings were evaluated on its test set as well as the test set of \textit{SemEval}. These datasets' classes represent events or topics. Interestingly, naive finetuning does not work for this application scenario. The training set contains a lot more classes than CrisisNLP and is therefore harder to train on; we assume that the embedding tends to overfit those classes and then cannot generalize to the novel ones in the test sets. However, the Siamese approach improves results in both cases, further confirming its generalizability. The embedding trained on all data sets performs worse on the \textit{Events2012} data, but better on \textit{SemEval}. This is probably the case because of the slight application shift in this data set, and the fact that the more similar \textit{SemEval} training set is used for training the ALL model (albeit with a different class selection).

\begin{figure}[htbp]
    \centering
    \begin{subfigure}[b]{0.5\textwidth}
        \centering
        \includegraphics[width=\textwidth]{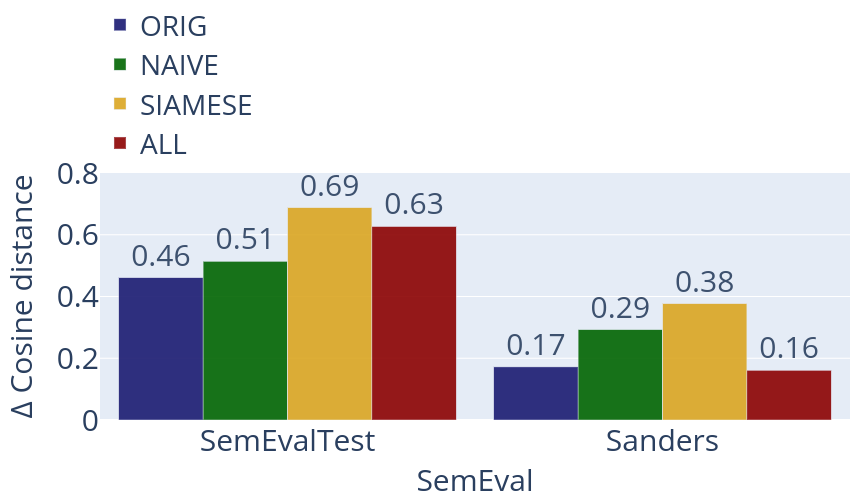}
        \caption{Experiment C}\label{fig:resC}
    \end{subfigure}%
    \begin{subfigure}[b]{0.5\textwidth}
        \centering
        \includegraphics[width=\textwidth]{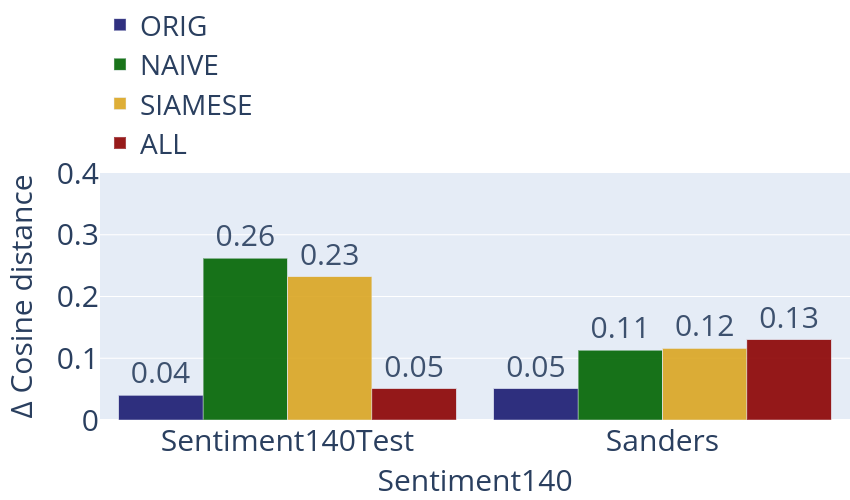}
        \caption{Experiment D}\label{fig:resD}
    \end{subfigure}%
    \caption{Experimental results}
    \label{fig:resCD}
\end{figure}

\subsection{Experiment C: SemEval $\rightarrow$ SemEvalTest \& Sanders }
Figure \ref{fig:resC} displays the results for another topic classification experiment. Naive and Siamese models were finetuned on the topic annotations of the \textit{SemEval} training set, and all models were tested on its test set and on the \textit{Sanders} data set. We once again see slight to medium improvements with the naive finetuning, and higher ones with the Siamese approach, confirming its better generalization ability. This is particularly interesting on the \textit{Sanders} test set because this data set is focused on just four unseen and semantically similar technology topics. Consequently, we see a lower performance with the ALL embedding; presumably, this happens because this very specific selection of topic becomes grouped closer together in the latent space. 

\subsection{Experiment D: Sentiment140 $\rightarrow$ Sentiment140Test \& Sanders}

\begin{figure}[htbp]
    \centering
    \begin{subfigure}[b]{0.33\textwidth}
        \centering
        \includegraphics[width=\textwidth]{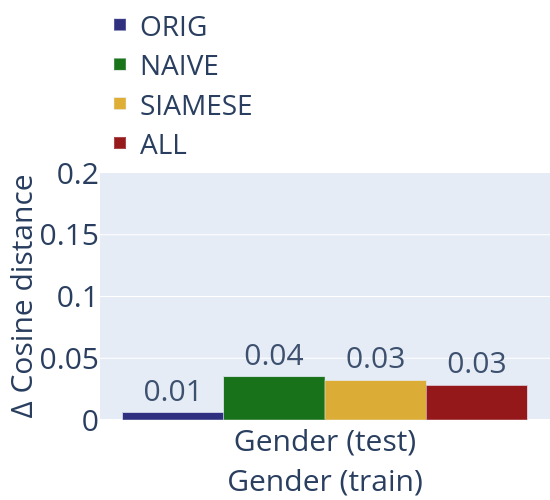}
        \caption{Experiment E}\label{fig:resE}
    \end{subfigure}%
    \begin{subfigure}[b]{0.33\textwidth}
        \centering
        \includegraphics[width=\textwidth]{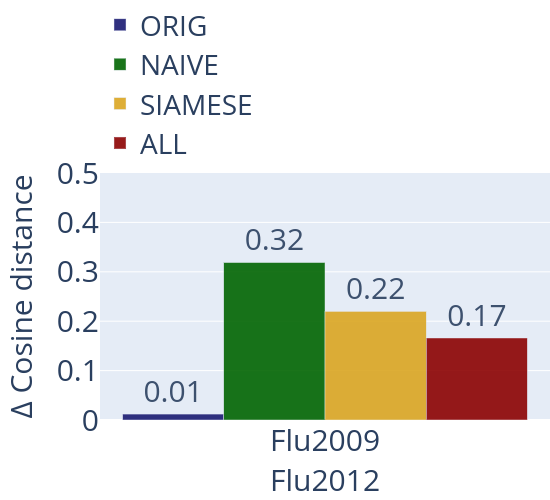}
        \caption{Experiment F}\label{fig:resF}
    \end{subfigure}%
    \begin{subfigure}[b]{0.33\textwidth}
        \centering
        \includegraphics[width=\textwidth]{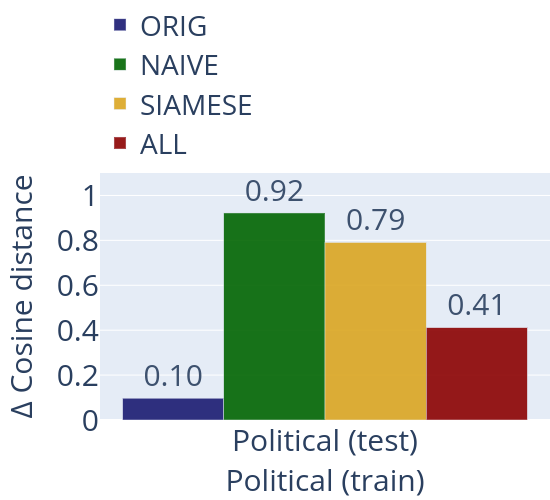}
        \caption{Experiment G}\label{fig:resG}
    \end{subfigure}%
    \caption{Experimental results}
    \label{fig:resEFG}
\end{figure}

Moving on from event and topic classification, figure \ref{fig:resD} shows the results for the sentiment recognition task. \textit{Sentiment140} is used for training here, and its test set as well as the sentiment annotations from \textit{Sanders} are used for evaluation. The naive finetuning displays a clear improvement over the USE baseline, demonstrating that this task is far enough removed to require finetuning. The Siamese approach performs slightly worse to equal; considering that the classes in this scenario do not change, it might not provide an advantage. For the \textit{Sanders}, the annotated classes are slightly different and the Siamese model may provide some flexibility there. The task is underrepresented in the model finetuned on all data, but this model still improves results on the unrelated \textit{Sanders} data set, perhaps covering the contained topics better.

\subsection{Experiment E: Gender}
Figure \ref{fig:resE} shows results for the gender classification task. As described above, this is a difficult task in general as there is not sufficient information in most tweets to discern the author's gender. As such, results are low. Finetuning in any way slightly improves them.

\subsection{Experiment F: Flu}
The last two experiments have a somewhat different setup because they deal with one predefined topic versus anything else, rather than the semantic space convered in the others. Figure \ref{fig:resF} shows the results for finetuning models to distinguish between tweets related to the seasonal flu vs. others. These others often still contain references to other types of flu, leading to the generally low results. Naive finetuning works best here; the two classes are known in advance, so the Siamese approach is not required for generalization.

\subsection{Experiment G: Political}
The last experiment is similar, but with an easier setup: Classifying political tweets versus any others. Naive finetuning improves results very much, whereas Siamese finetuning still helps a lot, but not quite as much. The ALL configuration delivers lower results (but still much higher than the baseline). This makes sense because a large amount of other training examples will lead to tweets on one topic being pushed closer together in the latent space.

\section{Conclusion}
In this paper, we present a novel method of finetuning sentence embeddings to specific domains and tasks. This method is based on a Siamese architecture which does not depend on the exact labels of a classification task. For this reason, the finetuned latent space can potentially generalize better to the new domain rather than specific classes, leading to a flexible embedding for future use cases in the domain. Our results on various combinations of data sets show that a naive finetuning improves results when compared to the baseline USE model, and the novel Siamese method is in many cases even better. This is particularly the case when testing on completely different datasets that do not concern the exact same task as the training model, confirming the Siamese approach's better generalization ability. As an exception, these models do not offer an advantage for binary classification tasks where the classes remain the same during evaluation.\\
Another effect of the Siamese approach is the ability to finetune on multiple data sets and tasks in parallel. Our results show that this often leads to even better results where knowledge from multiple sources can be exploited. Even when this is not the case, these embeddings perform significantly better than the baseline for almost all tasks, demonstrating their high flexibility.

\section{Future work}
There are many ways in which this approach could be extended for other tasks, types of data, or models. So far, we demonstrated usage on Twitter data sets due to their ability and the opportunity to compare various tasks on similar data. However, we see no reason why this approach would not work for other types of text data. In addition, we only tested performance on the Universal Sentence Encoder, but other sentence embeddings based on deep networks could be finetuned with the same paradigm. So far, only an English-language model was tested and only English-language data was used for finetuning. In a next step, we will also try this on the multilingual version of USE. Conceivably, merely using English-language data for finetuning a multilingual model may already lead to improved results in other languages.\\
Another direction we are currently exploring is a more detailed finetuning on one specific domain. Preliminary results already show that finetuning on multiple tasks related to disaster tweets in parallel can lead to improved results for all of them, e.g. recognition of disaster information, semantic classification, importance rating etc.\\
Finally, the Siamese approach can be adapted to other types of available data. So far, we considered text data sets with class labels. The same method could also be used for regression tasks by providing continuous instead of binary labels to compare the Cosine similarity against. A particularly interesting further development would be unsupervised finetuning, e.g. by assuming that consecutive sentences or tweets by the same user are often similar to each other.

\begin{small}
\bibliographystyle{coling}
\bibliography{coling2020}
\end{small}

\end{document}